\documentclass[sigconf]{acmart}

\usepackage{hyperref} 
\usepackage{hyperxmp} 
\usepackage{float}
\usepackage[ruled]{algorithm2e}
\usepackage{booktabs}
\usepackage{multirow}
\usepackage[autolanguage]{numprint}
\usepackage{paralist}
\usepackage{placeins}
\usepackage{caption}

\usepackage{amsmath,amssymb,amsfonts}
\usepackage{algorithmic}
\usepackage{graphicx}
\usepackage{textcomp}
\usepackage{xcolor}

\usepackage{array}
\usepackage{setspace}
\newcolumntype{L}[1]{>{\raggedright\let\newline\\\arraybackslash\hspace{0pt}}m{#1}}
\newcolumntype{C}[1]{>{\centering\let\newline\\\arraybackslash\hspace{0pt}}m{#1}}
\newcolumntype{R}[1]{>{\raggedleft\let\newline\\\arraybackslash\hspace{0pt}}m{#1}}
\newcolumntype{Z}[1]{>{\let\newline\\\arraybackslash\hspace{0pt}}m{#1}}
\newcolumntype{P}[1]{>{\centering\arraybackslash}p{#1}}

\def\h{\mathbf{h}}
\def\L{{\mathcal L}}
\def\Lc{{\mathcal L}_{\mathrm{SSL}}}
\def\Lcoles{{\mathcal L}_{\mathrm{CoLES}}}
\def\Lg{{\mathcal L}_{\mathrm{GNN}}}
\def\Llp{{\mathcal L}_{\mathrm{LP}}}
\def\Lwp{{\mathcal L}_{\mathrm{WP}}}

\def\Ls{{\mathcal L}_{\mathrm{sim}}}
\def\Lbpr{{\mathcal L}_{\mathrm{BPR}}}

\AtBeginDocument{%
  }

\copyrightyear{2026}
\acmYear{2026}
\setcopyright{cc}
\setcctype{by}
\acmConference[WWW '26]{Proceedings of the ACM Web Conference 2026}{April 13--17, 2026}{Dubai, United Arab Emirates}
\acmBooktitle{Proceedings of the ACM Web Conference 2026 (WWW '26), April 13--17, 2026, Dubai, United Arab Emirates}
\acmPrice{}
\acmDOI{10.1145/3774904.3792886}
\acmISBN{979-8-4007-2307-0/2026/04}

\begin{document}

\title{Beyond Isolated Clients: Integrating Graph-Based Embeddings into Event Sequence Models}

\author{Harry Proshian}
\affiliation{%
  \institution{Steklov Institute of Mathematics}
    \city{Saint Petersburg}
  \country{Russia}}

\author{Nikita Severin}
\affiliation{%
  \institution{Independent Researcher}
  \city{Belgrade}
  \country{Serbia}}

\author{Sergey Nikolenko}
\affiliation{%
  \institution{ISP RAS}
  \institution{Steklov Institute of Mathematics}
  \city{Saint Petersburg}
  \country{Russia}}

\author{Ivan Kireev}
\affiliation{%
  \institution{Sber AI Lab}
  \city{Moscow}
  \country{Russia}}

\author{Andrey Savchenko}
\affiliation{%
  \institution{Sber AI Lab}
  \institution{HSE University; ISP RAS}
    \city{Moscow}
  \country{Russia}}

\author{Ivan Sergeev}
\author{Maria Postnova}
\affiliation{%
  \institution{Sber}
  \city{Moscow}
  \country{Russia}}

\author{Ilya Makarov}
\affiliation{%
  \institution{AIRI; ISP RAS }
  \city{Moscow}
  \country{Russia}}

\renewcommand{\shortauthors}{Harry Proshian et al.}

\begin{abstract}
    Large-scale digital platforms generate billions of timestamped user-item interactions (events) that are crucial for predicting user attributes in, e.g., fraud prevention and recommendations. While self-supervised learning (SSL) effectively models the temporal order of events, it typically overlooks the global structure of the user-item interaction graph. To bridge this gap, we propose three model-agnostic strategies for integrating this structural information into contrastive SSL: enriching event embeddings, aligning client representations with graph embeddings, and adding a structural pretext task. Experiments on four financial and e-commerce datasets demonstrate that our approach consistently improves the accuracy (up to a 2.3\% AUC) and reveals that graph density is a key factor in selecting the optimal integration strategy.
    \end{abstract}



\maketitle

\section{Introduction}

Digital platforms generate billions of timestamped interaction events daily, including purchases, clicks, and transactions, forming temporal sequences that encode user behavior. Converting these raw streams into accurate predictions of entity attributes (i.e., who a customer is) drives product recommendations, credit risk management, fraud prevention, and personalized marketing. Even a modest 1\% improvement in fraud prediction AUC can translate into millions of dollars in revenue for a large institution.

Supervised learning (SL) was initially applied to event sequences~\cite{babaev2019rnn,alaraj22}, but obtaining labels for many properties (e.g., creditworthiness or fraud likelihood) requires months of observation and validation, limiting SL applicability. Self-supervised learning (SSL) addresses this by learning rich representations from unlabeled data~\cite{zhang2023contrastive} through pretext tasks including masked element recovery~\cite{padhi2021tabular,wang2024pretext}, contrastive learning~\cite{babaev2022coles,zbontar2021barlow}, or generative modeling~\cite{bazarova2025learning,oord2018representation}.

\begin{figure}
\includegraphics[width=\linewidth]{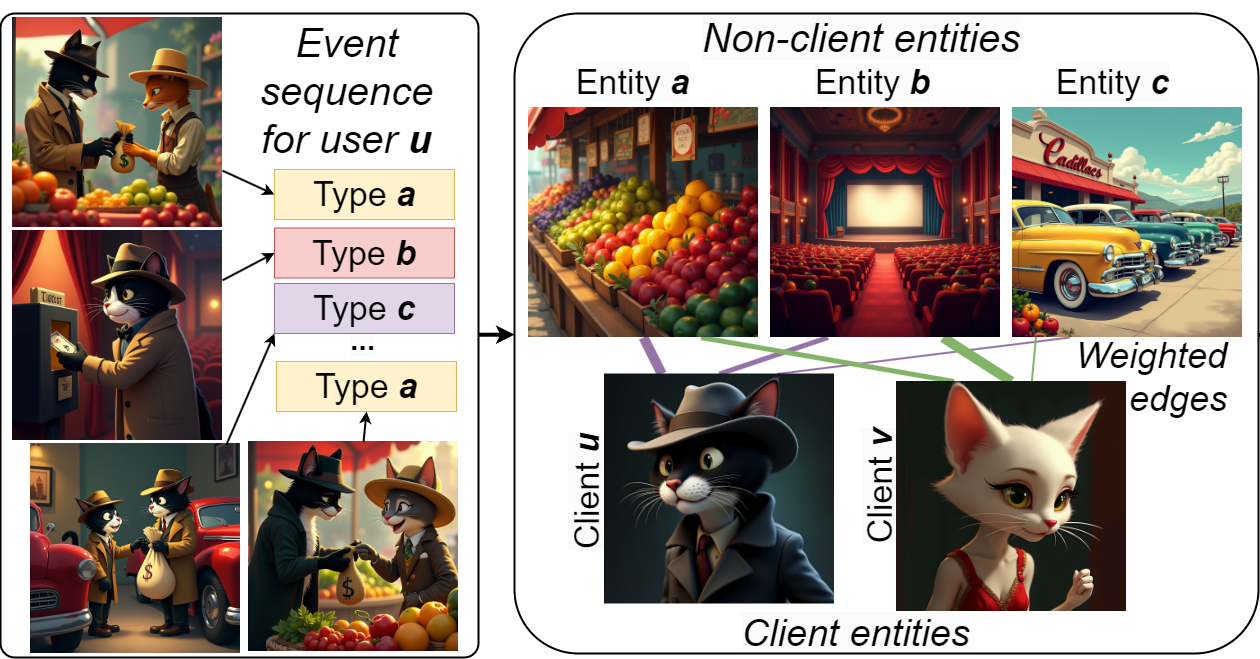}
\caption{Global interaction graph from event sequences}\label{fig:graph}
\end{figure}

However, existing SSL methods model each client independently, considering only their attributed event sequences. In practice, interactions with non-clients (items, e.g. products or posts in social networks) define a bipartite graph (see Fig.~\ref{fig:graph}) whose structure reflects global patterns useful for predictions but typically ignored.

\textbf{Our contributions.} We bridge this gap by integrating this structural information into contrastive SSL methods for event sequences (Fig.~\ref{fig:graph}). We propose three model-agnostic strategies: enriching event embeddings with graph features, aligning client representations with graph embeddings, and adding structural pretext tasks. Our key insight is that \emph{graph density determines the optimal integration strategy}: pretrained GNN embeddings excel on moderately dense graphs, while auxiliary losses remain robust across density extremes. Experiments on four datasets from financial and e-commerce domains, using CoLES~\cite{babaev2022coles} and Barlow Twins~\cite{zbontar2021barlow} SSL methods, show consistent gains that remain stable across two orders of magnitude in scale.
The source code of all our models and experiments are publicly available\footnote{\url{https://github.com/sb-ai-lab/WWW26_Graph-Based-Embeddings-for-Event-Sequences}}.

\section{Base SSL models for Event Sequences}\label{sec:base}

\def\bu{\mathbf{u}}
\def\bv{\mathbf{v}}

We adopt two popular contrastive SSL paradigms as the backbone for our graph-augmented framework: {CoLES} and {Barlow Twins}. Both methods share two low-level components: an \emph{event encoder} (maps a single interaction to a vector) and a \emph{sequence encoder} (aggregates a subsequence into an embedding), while differ in pair sampling and loss design.
For clarity, we recap each paradigm in its vanilla form and then, in Section~\ref{sec:method}, demonstrate how graph-based features can be integrated without altering the core architecture.

\textbf{CoLES}~\cite{babaev2022coles} 
has four components: an encoder, a subsequence sampling algorithm, a hard negative selection algorithm, and a margin-based contrastive loss function.

\emph{Subsequence sampling}.
Given a client sequence $\mathcal S$, CoLES draws $k$ random \emph{slices} (we set $k=2$) by choosing a length $L$ and a start index $s$ uniformly at random, preserving temporal order. This strategy yields $k-1$ positive counterparts for every anchor within a batch.

\emph{Encoding}.
Each event is mapped to a dense vector; a GRU~\cite{cho2014learning} then summarises the slice into a fixed-size embedding. We denote two such embeddings by $\bu$ and $\bv$.

\emph{Hard negative mining}.
The five closest (in Euclidean space) embeddings from \emph{different clients} within a batch are selected as negative samples, focusing the model on the most confusing examples.

\emph{Loss function}.
The final contrastive loss function for CoLES is
$$\Lcoles = Y_{\bu\bv} d(\bu,\bv)^{2} + \left(1 - Y_{\bu\bv}\right)
          \max\left\{0, \rho - d(\bu,\bv)\right\}^{2},$$
where $\bu$ and $\bv$ are two client (subsequence) embeddings, $Y_{\bu\bv}=1$ if $\bu$ and $\bv$ originate from the \emph{same} client, $d(\cdot,\cdot)$ is the Euclidean distance, and the margin $\rho$ prevents representations of different clients from collapsing.

\textbf{Barlow Twins} (BT)~\cite{zbontar2021barlow} 
BT avoids explicit negatives by maximizing similarity between two augmented views while decorrelating embedding dimensions (redundancy reduction).

\emph{Batch cross-correlation}.
Let $Z_A,Z_B \in {\mathbb R}^{N\times d}$ be the matrices of embeddings obtained from two random slices of the same $N$ clients.  
The empirical cross-correlation is
$
C = \frac{1}{N} Z_A^\top Z_B.
$

\emph{Loss function}.
The BT objective splits into the \emph{invariance} term
${\mathcal L}_{\mathrm{Inv}}=\sum_i (1-C_{ii})^2$  
and the \emph{redundancy} term
${\mathcal L}_{\mathrm{Red}}=\sum_{i\neq j}C_{ij}^2$, combined as
$
{\mathcal L}_{\mathrm{BT}} = {\mathcal L}_{\mathrm{Inv}} + \lambda{\mathcal L}_{\mathrm{Red}},
$
with $\lambda$ controlling the trade-off between them. 
No explicit negatives are required, which eliminates the computational overhead of hard negative mining present in CoLES and makes BT attractive for large-batch training.

\def\Ncl{N_{\mathrm{cl}}}
\def\Nncl{N_{\mathrm{ncl}}}

\section{Proposed Method}\label{sec:method}

We aim to enhance sequence-based SSL methods with \emph{global} relational context naturally encoded in interaction graphs. Our approach consists of three main steps.

\textbf{Graph Construction}. Given timestamped events (\emph{client}, \emph{non-client}, \emph{attributes}), we construct an undirected weighted bipartite graph
$G=(V_{\text{cl}} \cup V_{\text{ncl}},E)$ whose edges coincide with historical interactions (Fig. \ref{fig:graph}). The weight of an edge between client $u$ and non-client $v$ is proportional to their interaction count, rescaled by the inverse popularity of $v$ to limit the dominance of hub nodes.

\textbf{Feature extraction}. We experiment with two increasingly expressive feature extraction strategies for the graphs:
\begin{inparaenum}[(1)] 
\item \emph{adjacency matrix}, the raw $0$--$1$ or weighted incidence vector that captures first-order connections between entities;
\item \emph{graph neural networks (GNNs)} based on message passing mechanism, particularly: GCN~\cite{kipf2017semisupervised}, GraphSAGE~\cite{10.5555/3294771.3294869} and GAT~\cite{velickovic2018graph}.  
\end{inparaenum}
To mitigate over-smoothing~\cite{platonov2023critical}, we incorporate residual connections~\cite{resnet} between layers.

We evaluated all GNN variants but report main results using \emph{GraphSAGE} embeddings, which consistently delivered the best performance. Its effectiveness stems from robust generalization and superior scalability on large graphs, enabled by an efficient neighbor sampling strategy.

 \textbf{Integration into SSL}. We introduce three strategies to incorporate graph-based embeddings into any sequence-level SSL objective.

1. \textsc{GrEmb}: \emph{replacing non-client entity embedding layers with GNN embeddings}. GNN embeddings for non-client entities are used in SSL through joint or sequential training. In joint training, both are optimized simultaneously with a combined loss function $\L = \gamma \Lc + (1-\gamma)\Lg$. The graph-based loss function is defined as $\Lg = \alpha\Llp + (1-\alpha)\Lwp$,
where $\Llp$ is the binary cross-entropy (BCE) for link prediction, and $\Lwp$ is either BCE or mean squared error (MSE) for weight prediction. In sequential training, graph features are pretrained and used to initialize CoLES.

2. \textsc{Reg}: \emph{SSL regularization with trainable client GNN embeddings} that {aligns} client embeddings from two sources --- SSL model and GNN --- by passing graph-based client embeddings with SSL embeddings into the loss function $\Lc$. For each client anchor embedding, we add another positive example derived from the corresponding graph embedding, and hard negatives can come from either graph-based or sequence-based embeddings.

3. \textsc{Loss}: \emph{auxiliary loss function}. In this case, we preserve the entire SSL pipeline but introduce an additional pretext task that maintains relative positions of entities within their neighborhoods in graph. Before training SSL model, we use adjacency matrix embeddings (since they preserve the original structure) to calculate the cosine similarity between all clients. Then, for each anchor client in SSL loss we add a ranking loss $\Ls$ that ensures that its similarity ordering to other clients is preserved the same as from graph-based perspective. The resulting loss function is 
$\L = \gamma \cdot \Lc + (1 - \gamma) \cdot \Ls$, and $\Ls$ can be either the triplet loss or BPR
loss ~\cite{10.5555/1795114.1795167}: \begin{equation}
\Lbpr = - \sum\nolimits_{(u, v^+, v^-)} \log(\sigma(\h_u \cdot (\h_{v^+} - \h_{v^-}))),
\end{equation}
where $\h_u$, $\h_{v^+}$, and $\h_{v^-}$ are the anchor, positive, and negative client embeddings, respectively.

To mine positive and negative samples, we use a binning strategy that moves beyond random pairing. For each client, we precompute $m$ bins of other clients based on cosine similarity. During training, two bins are sampled from the current mini-batch, with the higher-similarity bin providing positives and the lower-similarity bin negatives, yielding more informative contrastive pairs. The number of bins and sampling repetitions are key hyperparameters.

\section{Experimental Evaluation}
\label{sec:eval}

\begin{table}[!t]\setlength{\tabcolsep}{6pt}\centering
\caption{Dataset statistics.}\label{tbl:datasets}
\begin{tabular}{lp{.1\linewidth}p{.1\linewidth}p{.1\linewidth}p{.1\linewidth}}\toprule
 & \textbf{Gen\-der} & \textbf{Age} & \textbf{MTS-ML-Cup} & \textbf{Inter\-nal} \\\midrule
Client entities              & \numprint{15}K & \numprint{50}K & \numprint{415}K & 1M  \\
Non-client entities          & \numprint{184} &    \numprint{204} &   \numprint{197}K  & 418 \\
Interactions                 & \numprint{6.85}M & \numprint{44}M & \numprint{323}M & 530M \\
Labeled clients              & \numprint{8.4}K &  \numprint{30}K & \numprint{270}K & 1M  \\
$\quad$training set  & \numprint{7.56}K &  \numprint{27}K & \numprint{243}K & 900K  \\
$\quad$test set   & \numprint{840} &    {3000} & {27}K  & 100K  \\
Training graph nodes      & \numprint{14.3}K & \numprint{47.2}K & \numprint{585}K & 900K  \\
Training graph edges     & \numprint{288}K & \numprint{1.97}M & \numprint{2.97}M & 42.3M \\
Graph density & .110 & .205 & .0004 & .112
\\\bottomrule
\end{tabular}
\end{table}

\textbf{Datasets}. 
Table~\ref{tbl:datasets} shows four datasets spanning two orders of magnitude in scale and three orders in graph density ($|E|/(|V_\mathrm{cl}|,|V_\mathrm{ncl}|)$):
\begin{inparaenum}[(1)]
    \item \emph{Gender}\footnote{\url{https://storage.yandexcloud.net/di-datasets/trans-gender-2019.zip}}, 
    a small dataset of banking transactions to predict a client's gender (binary classification);
\item \emph{Age}\footnote{\url{https://ods.ai/competitions/sberbank-sirius-lesson}}, 
a medium-sized banking transactions dataset to predict the clients' age group (multi-class classification);
\item \emph{MTS-ML-Cup}\footnote{\url{https://ods.ai/competitions/mtsmlcup}}, 
a large dataset of clients visiting web resources, where the task is to predict both the gender and age group;
\item \emph{Internal}, a private financial dataset from a large bank with $>$530M financial operations over 9 months; nodes represent clients and product categories, edges, financial operations.
\end{inparaenum}

\textbf{Experimental setup}. We evaluate the impact of integrating graph-based features into CoLES and Barlow Twins under different training strategies and configurations. For downstream tasks, we apply LightGBM \cite{NIPS2017_6449f44a} to the learned client embeddings and report AUC-ROC (AUC) and accuracy (Acc).

\begin{table}[!t]\centering
\setlength{\tabcolsep}{3pt}
\caption{Best experimental results.}
    \label{tbl:main_results_upd}
\begin{tabular}{llccccc}
        \toprule
        \multicolumn{2}{c}{\textbf{Method}} & \multicolumn{2}{c}{\textbf{Gender}} & \textbf{Age} & \multicolumn{2}{c}{\textbf{Internal}} \\
             \textbf{Base} & \textbf{Ours} & \textbf{AUC} & \textbf{Acc} & \textbf{Acc} & \textbf{AUC} & \textbf{Acc} \\
        \midrule
        Popular class & & 0.558 & 0.558 & 0.252 & 0.333 & 0.333 \\
        CoLES & & 0.877 & 0.793 & 0.637 & 0.905 & 0.748 \\
    CPC & & 0.851 & 0.768 & 0.602 & 0.899 & 0.741  \\
    BT & & 0.865 & 0.779 & 0.634 & 0.895 & 0.734 \\
    \multicolumn{2}{l}{Tabformer} & 0.847 & 0.764 & 0.601 & 0.894 & 0.733  \\
GPT & & 0.799 & 0.713 & 0.589 & 0.884 & 0.719 \\
    \midrule
    Adj. matrix & & 0.875 & 0.793 & 0.252 & 0.827 & 0.645 \\
    CoLES & \textsc{GrEmb}$_{\mathrm{SAGE}}$ & \underline{0.886} & \underline{0.808} & \underline{0.638} & \underline{0.908} & \underline{0.754} \\
    CoLES & \textsc{GrEmb}$_{\mathrm{Adj}}$ & \textbf{0.889}  & \textbf{0.811} & \underline{0.638} & \textbf{0.910} & \textbf{0.757} \\
    CoLES & \textsc{Loss} & {0.883} & 0.800 & \textbf{0.640} & 0.900 & 0.740 \\
    BT & \textsc{GrEmb}$_{\mathrm{SAGE}}$ & 0.870 & 0.789 & 0.637 & 0.895 & 0.733 \\
    BT & \textsc{Loss} & 0.869 & 0.791 & 0.636 & 0.894 & 0.733 \\
        \bottomrule
   \end{tabular}
\end{table}

\textbf{Complementarity of graph- and sequence-based spaces}.
We first verify whether graph and sequence representations capture different information. Figure~\ref{fig:dissimilarity} shows Jaccard dissimilarity between $k$-nearest-neighbor sets from adjacency-matrix embeddings vs. CoLES embeddings on Gender. Average dissimilarity exceeds 60\% and remains high even for $k=1000$ (large neighborhoods), confirming that the two modalities encode fundamentally different relational patterns. This motivates our integration strategy.

\textbf{Ablation study}. Table~\ref{tbl:gender} compares GNN integration strategies on Gender. {Sequential training} (learn GNN embeddings, then initialize SSL) consistently outperforms joint and regularization approaches, which exhibit substantial overfitting to graph reconstruction, introducing noise that hinders SSL convergence. Training a GNN solely via SSL gradients (without graph loss) actually degrades performance below baseline, confirming that GNNs require appropriate graph-side supervision to provide useful signals. Therefore, all main results use \textsc{GrEmb} with sequential training.

\begin{table*}[!t]\centering\setlength{\tabcolsep}{3pt}
\begin{minipage}{0.34\textwidth}
    \caption{Graph-based features for CoLES on \emph{Gender}}\label{tbl:gender}
    \begin{tabular}{lcc}
        \toprule
        \textbf{Approach} & \textbf{AUC} & \textbf{Acc} \\
        \midrule
        Base CoLES & 0.877 & 0.793 \\
$\quad$+ \textsc{Reg} & 0.855 & 0.768 \\
$\quad$+ \textsc{Emb}$_{\mathrm{SAGE}}$, joint training: & & \\ 
{$\quad\quad$ with GNN loss} & \underline{0.881} & \underline{0.799} \\
{$\quad\quad$ without GNN loss} & 0.872 & 0.789 \\
$\quad$+ \textsc{Emb}$_{\mathrm{SAGE}}$, seq. training & \textbf{0.886} & \textbf{0.808} \\
        \bottomrule
    \end{tabular}

\end{minipage}$\quad$\begin{minipage}{0.32\textwidth}

    \captionof{table}{CoLES results on MTS-ML-Cup; $\mathrm{Acc}_\mathrm{G}$, $\mathrm{Acc}_\mathrm{A}$ - accuracies of gender and age prediction}
    \label{tbl:results_mts_only}

    \begin{tabular}{llcc}
        \toprule
        \multicolumn{2}{c}{\textbf{Method}} & \multicolumn{2}{c}{\textbf{MTS-ML-Cup}} \\
        \textbf{Base} & \textbf{Ours} & $\mathbf{Acc}_\mathrm{G}$ & $\mathbf{Acc}_\mathrm{A}$ \\
        \midrule
        Popular & & & \\
        class & & 0.512 & 0.322 \\
        CoLES & --- & \underline{0.588} & \underline{0.382} \\
        CoLES & \textsc{GrEmb}$_{\mathrm{SAGE}}$ & \textbf{0.590} & 0.380 \\
        CoLES & \textsc{Loss} & \textbf{0.590} & \textbf{0.391} \\
        \bottomrule
    \end{tabular}

\end{minipage}$\quad$\begin{minipage}{0.28\textwidth}

\captionof{table}{$\Ls$ ablation study}\label{tbl:ls_all_datasets}\vspace{-.3cm}

\setstretch{.88}

\begin{tabular}{lrrrc}
\toprule
\textbf{$\Ls$} & \textbf{$\footnotesize\Lc$} & \multicolumn{2}{c}{\textbf{Gender}} & \textbf{Age}   \\
& \textbf{ratio} & \textbf{AUC} & \textbf{Acc} & \textbf{Acc}  \\
\midrule
None & 1 & 0.877 & 0.793 & 0.637  \\
\midrule
BPR  & 0.01 & 0.878 & 0.793 & 0.637  \\
loss & 0.15 & \textbf{0.883} & \underline{0.796} & \underline{0.638}  \\
        & 0.5  & 0.880 & \textbf{0.800} & \textbf{0.640}   \\
         & 0.85 & \underline{0.881} & 0.794 & 0.636   \\
         & 0 & 0.648 & 0.620 & 0.483  \\     
\midrule
Triplet  & 0.15 & 0.729 & 0.676 & 0.495   \\
loss             & 0.5  & 0.749 & 0.691 & 0.493 \\
             & 0.85 & 0.809 & 0.738 & 0.580 \\\bottomrule
\end{tabular}
\end{minipage}

\end{table*}

\begin{figure}[!t]\centering\setlength{\tabcolsep}{5pt}
    \centering
    \includegraphics[width=\linewidth]{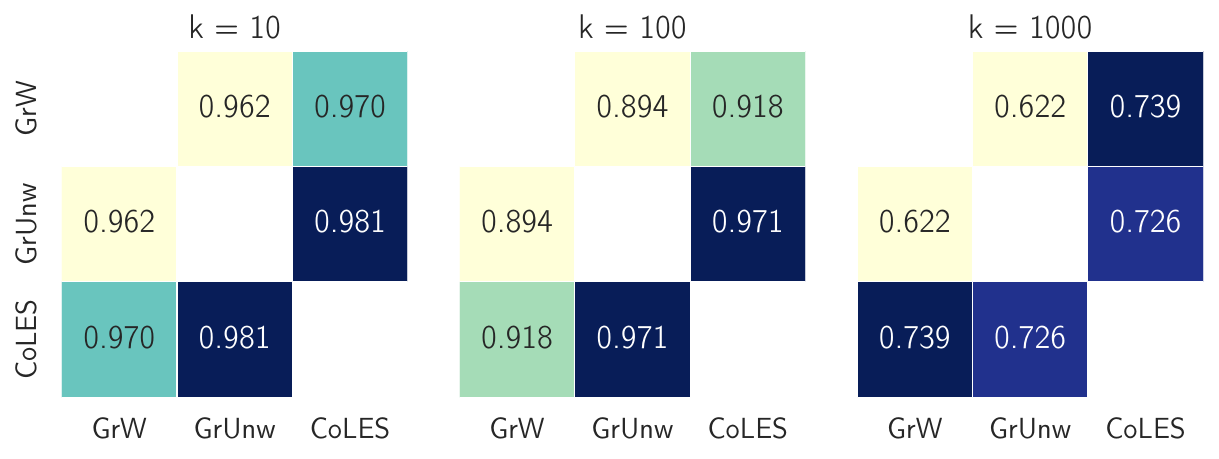}
    \caption{Latent space dissimilarity scores; GrW/GrUnw - features from the weighted/unweighted adjacency matrix.}
    \label{fig:dissimilarity}
\end{figure}

We also compared BPR-based ranking loss with standard triplet loss as an auxiliary objective (see Table ~\ref{tbl:ls_all_datasets}). BPR consistently outperforms triplet loss across all datasets, likely due to its more stable gradient properties and natural handling of implicit feedback. For $\gamma=0$ (pure graph loss) or high triplet loss weights we observed that the model fails to extract sequential patterns crucial for prediction, so balanced multi-objective learning is necessary.

\textbf{Main results}.
Tables~\ref{tbl:main_results_upd} (three banking datasets) and~\ref{tbl:results_mts_only} (MTS) summarize the primary experimental findings. Overall, integrating graph-based information consistently improves the performance of sequence-based SSL backbones, particularly on sparse and moderately dense graphs. Since MTS is a very large dataset with strict memory constraints, we evaluate only CoLES-based variants on it.

For the \emph{Gender} dataset, the best results were achieved by a hybrid approach. We first enhanced the CoLES model by initializing its non-client entity embeddings with pretrained GraphSAGE representations (the \textsc{GrEmb\textsubscript{SAGE}} method). After training CoLES, we concatenated learned client embeddings with the client's raw weighted and unweighted adjacency vectors before passing to the LightGBM classifier (\textsc{GrEmb\textsubscript{Adj}}). This strategy  improved CoLES by up to +1.3\% AUC and +2.27\% accuracy, showing graph-derived embeddings capture structural signals unavailable to sequence-only SSL objectives. Similar gains on the larger \emph{Internal} dataset with similar density confirm the robustness of this approach.

On datasets with extreme graph density, the behavior changes.
For dense \emph{Age}, adjacency-matrix embeddings collapse to almost random performance, and GNN-based \textsc{GrEmb} variants also lose effectiveness, which is consistent with classic over-smoothing effects in dense graphs. Here, the \textsc{Loss} approach yields the best results, with a slight but stable improvement, demonstrating that auxiliary alignment between graph and sequence views remains beneficial even when explicit GNN embeddings are not used. For \emph{MTS-ML-Cup}, the opposite extreme applies: the graph is very sparse, making structural signals weak at the node level. Here \textsc{Loss} is the strongest-performing variant too. This suggests that for extremely sparse graphs, learning a soft structural signal through the auxiliary objective is more effective than relying on explicit GNN embeddings.

In summary, these results confirm that the graph view is indeed complementary to sequence-based representations. GNN-derived embeddings provide strong improvements on average, but in extreme density regimes, learning a soft structural signal through the auxiliary loss is more effective than using explicit GNN embeddings.

\section{Conclusion}\label{sec:concl}

In this work, we have shown that incorporating bipartite graph structure into sequence-based SSL yields consistent improvements across four diverse financial and e-commerce datasets. We have revealed that \emph{graph density is the critical factor} determining optimal integration:
\begin{inparaenum}[(1)]
\item for moderate density (0.05--0.20), pretrained GNN embeddings (\textsc{GrEmb}) provide strong improvements by enriching event representations with structural context, while
\item for density extremes auxiliary similarity losses (\textsc{Loss}) remain robust when explicit GNN embeddings fail due to over-smoothing (dense) or noise (sparse).
\end{inparaenum}
The proposed framework is model-agnostic, adds minimal training overhead, and incurs zero inference cost.
Promising directions for future work include: 
\begin{inparaenum}[(1)]
\item dynamic graphs where edges evolve over time,
\item adaptive fusion with density-aware weighting of graph vs. sequence signals,
\item tighter coupling between sequence encoders and graph attention mechanisms,
\item theoretical analysis of the graph density regimes we identified empirically.
\end{inparaenum}

\section*{Acknowledgments} 
The work was supported by a grant, provided by the Ministry of Economic Development of the Russian Federation in accordance with the subsidy agreement (agreement identifier 000000C313925P4G0002) and the agreement with the Ivannikov Institute for System Programming of the Russian Academy of Sciences dated June 20, 2025 No. 139-15-2025-011.

\bibliographystyle{ACM-Reference-Format}
\bibliography{paper.bib}

\end{document}